# The Causal Topography of Cognition

Stevan Harnad
Chaire de recherche du Canada
Département de psychologie
Université du Québec à Montréal
Montréal, Québec
Canada H3C 3P8
http://www.crsc.uqam.ca/fr/index2.html
&
Electronics and Computer Science
University of Southampton
Highfield, Southampton
SO17 1BJ UNITED KINGDOM
http://users.ecs.soton.ac.uk/harnad/

**ABSTRACT:** *The causal structure of cognition can be simulated but not implemented computationally, just as the causal structure of a furnace can be simulated but not implemented computationally. Heating is a dynamical property, not a computational one. A computational simulation of a furnace cannot heat a real house (only a simulated house). It lacks the essential causal property of a furnace. This is obvious with computational furnaces. The only thing that allows us even to imagine that it is otherwise in the case of computational cognition is the fact that cognizing, unlike heating, is invisible (to everyone except the cognizer). Chalmers's "Dancing Qualia" Argument is hence invalid: Even if there could be a computational model of cognition that was behaviorally indistinguishable from a real, feeling cognizer, it would still be true that if, like heat, feeling is a dynamical property of the brain, a flip-flop from the presence to the absence of feeling would be undetectable anywhere along Chalmers's hypothetical component-swapping continuum from a human cognizer to a computational cognizer -- undetectable to everyone except the cognizer. But that would only be because the cognizer was locked into being incapable of doing anything to settle the matter simply because of Chalmers's premise of input/output indistinguishability. That is not a demonstration that cognition is computation; it is just the demonstation that you get out of a premise what you put into it. But even if the causal topography of feeling, hence of cognizing, is dynamic rather than just computational, the problem of explaining the causal role played by feeling itself – how and why we feel – in the generation of our behavioral capacity – how and why we can do what we can do – will remain a "hard" (and perhaps insoluble) problem.*

David Chalmers's (2012) thesis on cognition and computation is really two theses: (1) the "thesis of computational explanation"("computation provides a general framework for the explanation of cognitive processes"), which is true, but just the physical version of the Church/Turing Thesis (Piccinini 2011) according to which most physical processes are computable (to at least a close approximation); and (2) the "thesis of computational sufficiency" ("the right kind of computational structure suffices for the possession of a mind"), which, I will argue, is false (Harnad 1994).

**1. Computational Explanation.** The thesis that is true is that (just about) any causal system – whether it's a galaxy, a gall-bladder or a grain of sand -- can be modeled computationally, thereby (if successful) fully "capturing" (and hence explaining) its (relevant) causal mechanism -- in other words, explaining, causally, how it works.

This explanatory power is certainly something that psychology, cognitive science, neuroscience and artificial intelligence would want, for the kinds of things they study and hope to explain causally: organisms, their brains, and their behavior, as well as artificial devices we build that are capable of similar kinds of behavior. But it's an explanatory power that physics, chemistry, biology and engineering already have for the kinds of things that they study, without the need for "A Computational Foundation for the Study of Planetary Motion" (or of Helium, or of Hemoglobin, or of Convection Heaters). It's simply the ubiquitous observation that – alongside language and mathematics – computers and computational algorithms are useful tools in explaining the things there are in the world and how they work. (This is what Searle 1980 called "Weak AI.")

**2. Computational Sufficiency**. Chalmers's second thesis is that – unlike, say, flying, or digesting, which can likewise be modeled and explained computationally (*but are not, as Chalmers agrees, themselves instances of computation*) – cognition *is* (just) computation:

> *"[A] system implements a computation if the causal structure of the system mirrors the formal structure of the computation. The causal topology of a system… [is its] abstract causal organization…: the pattern of interaction among parts… abstracted away from the make-up of individual parts and from the way the causal connections are implemented… [A] property P [is] an organizational invariant if it is invariant with respect to causal topology: that is, if any change to the system that preserves the causal topology preserves P…"*

> *"Most properties are not organizational invariants. The property of flying is not … Digestion is not: if we gradually replace the parts involved in digestion with pieces of metal, while preserving causal patterns, after a while it will no longer be an instance of digestion… [M]ost properties depend essentially on certain features that are not features of causal topology. Flying depends on height, digestion depends on a particular physiochemical makeup… It is true that any given instance of digestion will implement some computation, as any physical system does, but the system's implementing this computation is in general irrelevant to its being an instance of digestion…*

> *"With cognition, by contrast, the claim is that it is in virtue of implementing some computation that a system is cognitive… mentality is an organizational invariant."*

I will try to flesh out both of Chalmers's theses without getting bogged down in technical details that are interesting but not pertinent to this fundamental distinction.

**3. Algorithms.** Computation is symbol manipulation: symbols are objects of arbitrary shape (e.g., 0 and 1) and they are manipulated on the basis of rules ("algorithms") that operate on the *shapes* of the symbols, not their *meanings*. In other words, computation is syntactic, not semantic. However, most computations can be interpreted as meaning something (otherwise we would not bother designing and doing them): We are interested in algorithms that can compute something useful, whether it's planetary motion or payroll checks.

How do computations do useful things? There are many ways. Numerical algorithms compute quantitative results we are interested in knowing. Desk calculators implement numerical algorithms. Boolean (and/or/not) search in Google's database can retrieve documents or data we are interested in. Virtual Reality can fool, entertain, or train our senses and movements. NASA's flight simulations anticipate problems that might arise in actual space flights. If Copernicus and Galileo had had digital computers, they might (just might!) have reached their conclusions faster, or more convincingly. Appel & Haken proved the four-color theorem with the help of computation in 1976. And if Mozart had had a computer to convert keyboard improvisation into metered notation, ready to print or edit and revise online, humankind might have been left a much larger legacy of immortal masterpieces from his tragically short 35 years of life.

**4. Causal Structure**. A word about causal structure – a tricky notion that goes to the heart of the matter. Consider gravitation. As currently understood, gravitation is a fundamental causal force of attraction between bodies, proportional to their respective masses. If ever there was a prototypical instance of causal structure – causing -- gravitational attraction is such an instance.

Now gravitational attraction can be modeled exactly, by differential equations, or computationally, with discrete approximations. Our solar system's planetary bodies and sun, including the causal structure of their gravitational interactions, can be "mirrored" formally in a computer simulation to as close an approximation as we like. But no one would imagine that that computer simulation actually embodied planetary motion: It would be evident that there was nothing actually moving in a planetary simulation, nor anything in it that was actually exerting gravitational attraction. (If the planetary simulation was being used to generate a Virtual Reality, just take your goggles off.)

The computer implementation of the algorithm would indeed have causal structure – a piece of computer hardware, computing, is, after all, a physical, dynamical system too, hence, like the solar system itself, governed by differential equations. But the causal structure of the implementation, as a dynamical system in its own right, would be the *wrong* causal structure (and would obey the wrong differential equations), insofar as planetary motion was concerned. It would not be the causal structure of planets, moving, nor gravity, attracting: it would be the causal structure of computer hardware, executing a certain algorithm, thereby formally "mirroring" the causal structure of planetary motion, as encoded in the algorithm. Two different dynamical systems, with different dynamical properties: those of the hardware, implementing the algorithm, and those of the planets, orbiting. (This is exactly the same as the two examples Chalmers concedes to be non-computational: flying and digestion.)

So in what sense does the causal structure of the computational model "mirror" the causal structure of the thing that it is modeling? The reply is that it mirrors it *formally*. That means that the symbols and symbol manipulations in the model can be systematically *interpreted* as having counterparts in the thing being modeled. (Virtual Reality can even make it into a visual interpretation, for our senses.) We don't even have to resort to computational simulations of planetary motion, nor even to the exact differential equations of physics in order to see this: We can see it in geometry, in the way $x^2 + y^2 = r^2$ "mirrors" the shape of a circle: Yes, $x^2 + y^2 = r^2$ "captures" the invariant structure of the circle, but it is not a circle, it is not shaped like a circle. (Reminder: a circle is the kind of thing you see on the Japanese flag!) No one would think otherwise, despite the accurate – indeed isomorphic -- mirroring. (And again, if the algorithm is dynamically generating something that looks round, like a circle, via VR, just take your goggles off.)

**5. Formally Mirroring Versus Physically Instantiating Causality.** But this is all obvious. Everyone knows that the mathematical (or verbal) description of a thing is not the same kind of thing as the thing itself, despite the formal invariance they share. Why would one even be tempted to think otherwise? We inescapably see *by observation* that a computational solar system lacks the essential feature of a real solar system despite the formally mirrored "causal structure," namely, there are no bodies there, *moving*, any more than there is anything *round* in the formal equation for a circle. The model – whether static or dynamic -- is just an explanatory device, a formal oracle, not a reincarnation of the thing it is modelling.

So it is evident by inspection in the case of physics, chemistry, and biology (e.g., when we note that synthetic hearts pump blood, but computational hearts do not), and even in mathematics, that the "causal structure" of the model (whether computational or analytic, symbolic or numeric, discrete or continuous, approximate or exact) may be the right one for a full causal *explanation* of the thing being modeled (explanation being a formal exercise), but not for a causal *instantiation* -- unless it actually embodies the relevant causal properties of the thing being modeled (the way a synthetic heart does) rather than just formally "mirroring" its properties by physically implementing their computation (the way a computational heart would do). Why is this not so evident in the case of cognition?

**6. Cognition Is Visible Only To the Cognizer.** How could Chalmers (and so many others) have fallen into the error of confusing these two senses of causality, one formal, the other physical? The reason is clear and simple (indeed Cartesian), even though it has been systematically overlooked. This very error is always made in the special case of cognition. What on earth is cognition? Unlike, say, movement, cognition is *invisible to all but the cognizer*! (1) We all know what cognizing organisms can do. (2) We all know what brain activity is. (3) And we all know *what it feels like to cognize*. Both behavioral capacity (1) and brain activity (2) are perfectly visible: they give rise to ordinary empirical data (observable, detectable, measurable). But we do not yet know what it is about the brain activity that generates (causes) the behavioral capacity, let alone the cognition. And the only one to whom *cognition itself* is "visible" is the cognizer (3). So when we theorize about what causes cognition (rather than just behavioral capacity), we are theorizing about something that is invisible to everyone except the cognizer, namely, cognizing itself: We are theorizing about what brain activity (or synthetic device activity, if cognizing is possible in synthetic devices) generates not only our observable behavior (and behavioral capacity) but also our unobservable cognizing.

**7. Cognizing Versus Moving.** Let's contrast the case of the brain and its invisible cognizing with the case of planets, moving, as well as with the case of a bodily organ other than the brain, one for which the problem of invisible properties does not arise: the heart, pumping. The reason we would never dream of saying that planetary motion was just computational -- or of saying that planets in a computational model were actually moving because the model "mirrors" their "causal structure" -- is simply the fact that planetary motion is *visible* (or observable by instruments). Hence it is inescapably obvious that the computational model, even if it shares the formal causal properties of planetary motion, *does not move*. The same is true of the computational heart: Unlike the synthetic heart, which really can pump blood (or some other liquid), a (purely) computational heart cannot pump a thing. (And note that I am talking about a physically implemented computational heart that faithfully mirrors the causal structure of a real heart -- formally, algorithmically, "isomorphically." It is merely symbolically pumping symbolic blood.)

**8. Bodily Activity, Brain Activity, and Cognizing.** Now imagine the same thing for the brain. It's a bit more complicated, because, unlike the heart, the brain is actually "doing" not one, nor two but *three* different things. One thing is generating (1) behavioral capacity: The brain is generating just about everything our bodies *do*, and are able to do, in the external world. The second thing is (2) the internal activity of the brain itself (the action potentials and secretions that are going on inside it). And finally, the brain is (3) cognizing (whatever that turns out to be – we will return to this).

So, unlike the planets and the heart, which are doing just one kind of thing, all of it fully observable to us (moving and pumping, respectively), the brain is doing *three* kinds of things, two of them visible (behavior and brain activity), one of them not (cognition).

**9. Turing Testing.** Now we are in a position to pinpoint exactly where the error keeps creeping in: No one would call a cognitive model a success if it could not be demonstrated to generate our behavioral capacity – if it could not do what we can do. So the question becomes: what kind of model can generate our behavioral capacity? That's where the Turing Test (TT) comes in (Harnad 2008): A model can generate our behavioral capacity if it can pass TT -- the full robotic version of TT, not just the verbal version (i.e., the ability to *do* everything a human can do in the world, not just to talk about it): A sensory-motor system that could pass the robotic TT would have to be able to perform indistinguishably from any of us, for a lifetime .

**10. "Neuralism."** Let's set aside the second kind of thing that brains do -- internal brain activity -- because it is controversial how many (and which) of the specific properties of brain activity are necessary either to generate our behavioral capacity or to generate cognition. It could conceivably turn out to be true that the only way to successfully generate cognition is one that preserves some of the dynamic (noncomputational) features of brain activity (electrochemical activity, secretions, chemistry etc.). In that case the fact that those observable (neural) features were missing from the computational model of cognition would be just as visible as the fact that motion was missing from the computational model of planetary motion (or that a computational plane was not flying, or a computational stomach not digesting). Let's call the hypothesis that, say, biochemical properties are essential – either to generate our behavioral capacity or to generate cognition -- "neuralism."

It's important to understand that my critique of the thesis that cognition is computation does *not* rest on the assumption that neuralism is true. We will discuss the issue of the implementation-independence of computation in a moment, but I don't want to insist that generating cognition depends on a requirement of neurosimilitude (necessarily preserving some of the dynamic properties of the brain). Neurosimilitude would definitely be needed in order to explain how the brain works, but not necessarily in order to explain either how to generate the brain's behavioral capacity or to explain how to generate cognition. Synthetic, non-neural means might be able to generate it too.

The basis of my critique of cognitive computationalism is, however, analogous to neuralism, and could perhaps be dubbed "dynamism." It is not that the brain's specific dynamic properties may be essential for cognizing, but that *some* dynamic properties may be essential for cognizing. We will return to this when we consider Chalmers's "Dancing Qualia" argument, according to which there could be a seamless transition from a real cognizing body and brain to a purely computational cognizer with all the same causal powers, replacing each internal component, one by one, by a purely computational simulation of it, the endpoint being the reconfiguration of a computer by its software so as to give it all the causal powers of a cognizing brain.

**11. Sensing and Doing.** Consider behavioral capacity first: Let us agree at once that whatever model we build that succeeds in generating our actual behavioral capacity – i.e., the power to pass the full robotic version of TT, for a lifetime – would definitely have *explained* our behavioral capacity (1), fully and causally, regardless of whether it did it via secretions or computations. If it were doing it computationally, by implementing an algorithm, it's clear that it would also need a robotic body, with sensors and moving parts, and that *those* could not be just the implementations of algorithms any more than flying or digestion could be.

Sensing, like moving (and flying, and digesting), is not implementation-independent symbol-manipulation. Consider the question of whether there could be a successful TT-passing robot that consisted of nothing other than (i) sensors and movable peripheral parts plus (ii) a functional "core" within which all the work (other than the I/O [sensory input and motor output] itself) was being done by an independent computational module that mirrored the causal structure of the brain (or of any other system capable of passing the TT). This really boils down to the question of whether the causal link-up between our sensory and motor systems, on the one hand, and the rest of our nervous system, on the other, can in fact be split into two autonomous modules -- a peripheral sensorimotor one that is necessarily noncomputational, plus a central one that is purely computational.

**12. Computational Core?** I cannot answer the question of whether such a causal split is possible or makes sense. To me it seems just as unlikely as the possibility that we could divide heart function (or flying, or digestion) into a noncomputational I/O module feeding into and out of a computational core. I think *sensorimotor function is mostly what the brain does through and through*, and that the intuition of a brain-in-a-vat receiving its I/O from the world – the intuition from which the computational-core-in-a-vat intuition derives -- is both unrealistic and homuncular.

But let us agree that the possibility of this functional partition is an empirical question*, depending on whatever will turn out to be the causal power it takes to generate our behavioral capacity*. If TT could not be passed by a computational core plus I/O peripherals then cognitive computationalism fails. But if TT could be successfully passed by a computational core plus I/O peripherals, would that entail that cognitive computationalism was correct after all?

**13. It Feels Like Something To Cognize.** This is the point to remind ourselves that we've left out the third burden of cognitive theory, in addition to (1) behavioral capacity and (2) brain function (which we've agreed to ignore): Even if we *define* cognition as whatever it takes to generate TT capacity, there is something the TT leaves out, something invisible to everyone except the cognizer, namely, consciousness: it *feels like something* to cognize (3). But that third property, unlike movement or secretions, cannot be perceived by anyone other than the cognizer himself. And I think it is this invisibility of cognition that is the real reason for Chalmers's error of confusing the computational implementation of the "causal structure" of cognition with the causal instantiation of cognizing itself: It looks from the outside as if there is no difference between what is going on inside a computational model of cognition and what is going on inside the brain of a cognizer. And it is for that reason that computation alone looks like a viable candidate for actually *instantiating*, rather than merely explaining cognition.

It is not that Chalmers is unaware of this distinction. He writes:

> *"[M]entality is an organizational invariant. [This] claim can be justified by dividing mental properties into two varieties: psychological properties - those that are characterized by their causal role, such as belief, learning, and perception - and phenomenal properties, or those that are characterized by the way in which they are consciously experienced. Psychological properties are concerned with the sort of thing the mind does, and phenomenal properties are concerned with the way it feels."*

**14. The "Dancing Qualia" Argument.** But Chalmers thinks his "dancing qualia" argument shows that feeling must, like computation itself, be an implementation-independent property, present in every implementation of the algorithm that successfully captures the right causal structure, no matter how radically the implementations differ:

> *"Assume conscious experience is not organizationally invariant. Then there exist systems with the same causal topology but different conscious experiences... we can (in principle) transform the first system into the second by making only gradual [dynamic] changes.... But given the assumptions, there is no way for the system to <u>notice</u>* [emphasis added] *these changes. Its causal topology stays constant, so that all of its functional states and behavioral dispositions stay fixed. If noticing is defined functionally (as it should be), then there is no room for any noticing to take place, and if it is not [defined functionally], any noticing here would seem to be a thin event indeed... it would be utterly impotent; it could lead to no change of processing within the system, which could not even mention it."*

What Chalmers is saying here is that if we hypothesize that there could be two physically different but "organizationally invariant" implementations of the same causal structure, one feeling one way and the other feeling another way (or not feeling at all), both variants implemented within the same hardware so that we could throw a switch to flip from one implementation variant to the other, the fact that the causal structure was the same for both variants would prevent the hypothetical difference in feeling from being felt. So the causal invariance would guarantee the feeling invariance.

But the trouble with this argument is that it *fails to take into account the fact that feeling, unlike movement, is invisible (to all but the feeler)*, yet, like movement, real.

**15. Implementation-Dependence Properties.** The reason the flip/flop thought experiment could not guarantee that all implementations of the causal structure of the solar system or the heart would move and beat, respectively, is that *moving and*

*pumping (and flying and digestion) are not computational properties*: We can *see* that. (It is empirically "observable.")

In the case of feeling, the reason this very same distinct possibility (that feeling is not a computational property) is not equally evident (as it ought to be) is that *the only one feeling the feeling* (or not-feeling the non-feeling, as the case may be) *is the cognizer* (or non-cognizer, not-feeling, in case there's no feeling going on, hence no cognition). But unless we are prepared to declare feeling to be identical with observable behavioral capacity *by definition* -- which would be tantamount to declaring that cognitive computationalism is true by definition – we have to allow for the possibility that feeling, like moving or flying, is not an implementation-independent computational property but a dynamical, *implementation-dependent* one.

For if feeling is (invisibly) like moving, then flipping from one causally invariant implementation to another could be like flipping between two causally invariant implementations of planetary motions, one that moves (because it really consists of planets, moving) and one that does not (because it's just a computer algorithm, encoding and impementing the same "organizational invariance"): In the case of feeling, however, unlike in the case of moving, the only one who would "see" this difference would be the cognizer, as the "movements" (behavior) would (*ex hypothesi*) have to be identical.

**16. Argument By Assumption.** Chalmers's "Dancing Qualia" argument simply does not take this distinct possibility into account at all. It won't do to say that there might be a felt difference, but it would have to be a "thin" one (because it could have no behavioral consquences). Chalmers gives no reason at all why the state difference could not be as "thick" as thick can be, as mighty as the difference between mental day and nonmental night, if the flip/flop were between a feeling and a non-feeling state, rather than just between two slightly different feeling states with a "thin" difference between them. But because Chalmers has imposed his premise of "functional" (i.e., empirical, observational) indistinguishability (with feeling unable to contradict that premise, because feeling is not externally observable), this would entail that *his premise itself* (not reality: Chalmers's premise) locks us into the consequence that even a flip from an insentient "Zombie" state to a behaviorally indistinguishable conscious state could make no empirically detectable difference. The feeling state says, truly, "I am feeling," and it is; the nonfeeling state says (as obliged by the premise, but falsely) "I am feeling," but it's not.

I am not saying that I believe there could be a behaviorally indistinguishable Zombie state (Harnad 1995) -- just that it cannot be ruled out simply on the grounds of having assumed a premise! (Or, better, one must not assume premises that could allow empirically indistinguishable Zombies.) Nor is it necessarily true that "noticing" has to be "functional" – if functional means computational: it *feels like something* to notice; and what is on trial here is *whether computationalism alone can implement feeling at all*, as opposed to merely simulating its externally observable correlates, formally. Chalmers's argument trades on the fact that behavioral indistinguishability entails that we cannot see any difference between noticing and not-noticing – feeling and not-feeling – except if it has some external behavioral (or internal computational) expression. But *who promised that the system could feel anything at all, if it was merely computational?* There remains the distinct possibility that what would turn the mental lights on or off would be a *dynamical property* – as in digesting, heating or flying -- rather than a computational one. (This is the "dynamism" mentioned earlier.)

"Demoting" feeling to a dynamical property, its presence or absence dependent on conformity with the right differential equations rather than the right computer program releases feeling from having to be an implementation-independent computational property.

**17. The Easy Problem.** But does cognition have to be *felt* cognition? Elsewhere, Chalmers (1995) has given a name to a longstanding distinction that – unlike the distinction between sensorimotor peripherals and computational core – marks a solid empirical difference: The problem of explaining behavioral capacity (doing) – i.e., how and why we can *do* what we can do -- is "easy" to solve. Cognitive science is nowhere near solving it, but there seem to be no principled obstacles. The problem of explaining how and why we feel (the "mind/body problem"), in contrast, is hard to solve, perhaps even impossible (Harnad 1995, 2000).

So there's something to be said for concentrating on solving the "easy" problem of explaining how to pass the TT first (Harnad & Scherzer 2008). I think it is unlikely (for reasons that go beyond the scope of this discussion) that the solution to even that "easy" problem will be a purely computational one (Harnad 1994). It is more likely that the causal mechanism that succeeds in passing TT will be hybrid dynamic/computational – and that the dynamics will not be those of the hardware implementation of the computation (those really *are* irrelevant to cognition). The dynamics will be those of the sensory-motor surface – plus all the internal topographic (spatial shape-preserving) dynamics in between (Silver & Kastner 2009), possibly including molecular shape dynamics as well. That – and not computation alone – will be the "causal topography" of cognition.

**18. The Hard Problem.** Once that ("easy") problem is solved, however, only the TT-passing system itself will know whether it really does cognize – i.e., whether it really feels like something to be that TT-passer. (Being able to determine *that* through empirical observation would require solving the (insoluble) other-minds problem.)

But even if we had a guarantee from the gods that the TT-passer really cognizes, that still would not explain the *causal role* of the fact that the TT-passer feels. The causal role of feeling in the causal topography of cognition will continue to defy explanation even if feeling really is going on in there -- probably because there is no more explanatory "room" left in a causal explanation (not just "thin" room, but no room), once all the relevant dynamics are taken into account (Harnad 2011).

So not only is it unlikely that implementing the right computations will generate cognition, but whatever it is that does generate cognition -- whether its causal topography is computational, dynamical, or a hybrid combination of both – will not explain the causal role of consciousness itself (i.e., feeling), in cognition. And that problem may not just be "hard," but insoluble.[1]

---

[1] If it were true, *neuralism* – the hypothesis that certain dynamical properties of the brain are essential in order to generate the brain's full behavioral capacity -- would already refute *computationalism*, because it would make the premise that computation alone could generate the brain's full causal power false. The truth of neuralism would also refute Chalmers's "Dancing Qualia," argument, *a fortiori*. Even as a possibility, neuralism invalidates the Dancing Qualia argument. *Dynamism* provides a more general refutation than neuralism; it does not depend on the brain's specific dynamics having to be the only possible way to generate either the brain's behavioral capacity or feeling. It is based on the fact that feeling, like moving, is a dynamical property, but, unlike moving, observable only to the feeler. Hence both computationalism and the Dancing Qualia argument could fail, with no one able to detect their failure except the feeler -- while the computationalist premise pre-emptively defines the feeler *a priori* as unable to signal the failure of the premise in any way. This is not just a symptom of the circularity of Chalmers's premise about the causal power of computation, however. It is also a symptom of the "hard problem" of explaining the causal role of feeling: Except if (contrary to all empirical evidence to date) dualism were true -- with feeling being an autonomous psychokinetic causal force, on a par with gravitation and electromagnetism -- feeling seems to be causally superfluous in any explanation of cognitive capacity and function, regardless of whether the causality is computational, dynamic or hybrid.